\documentclass{article}
\usepackage{ijcai18}
\usepackage{times}
\usepackage{xcolor}
\usepackage{soul}
\usepackage[utf8]{inputenc}

\usepackage{balance}

\usepackage[small]{caption}
\usepackage{amsmath,graphicx}
\usepackage{subcaption}
\usepackage{multirow}

\usepackage{ dsfont }
\usepackage{amsthm}

\def\name{DRoP}

\usepackage[linesnumbered,ruled,vlined,algo2e]{algorithm2e}

\title{Interactive Reinforcement Learning with Dynamic Reuse of Prior Knowledge}

\author{
Zhaodong Wang\\
School of EECS\\
Washington State University\\
zhaodong.wang@wsu.edu
\And
Matthew E. Taylor\\
School of EECS\\
Washington State University\\
taylorm@eecs.wsu.edu
}

\begin{document}

\maketitle

\begin{abstract}
Reinforcement learning has enjoyed multiple successes in recent years. However, these successes typically require very large amounts of data before an agent achieves acceptable performance. This paper introduces a novel way of combating such requirements by leveraging existing (human or agent) knowledge. In particular, this paper uses demonstrations from agents and humans, allowing an untrained agent to quickly achieve high performance. We empirically compare with, and highlight the weakness of, HAT and CHAT, methods of transferring knowledge from a source agent/human to a target agent. This paper introduces an effective transfer approach, \name{}, combining the offline knowledge (demonstrations recorded before learning) with online confidence-based performance analysis. \name{} dynamically involves the demonstrator's knowledge, integrating it into the reinforcement learning agent's online learning loop to achieve efficient and robust learning.

\end{abstract}

\section{Introduction}








There have been increasingly successful applications of reinforcement learning~\cite{sutton1998reinforcement} methods in both virtual agents and physical robots. 
In complex domains, reinforcement learning (RL) often suffers from slow learning speeds, which is particularly detrimental when initial performance is critical. External knowledge may be leveraged by RL agents to improve learning --- demonstrations have been shown to be useful for many types of agents' learning~\cite{schaal1997learning,argall2009survey}. To leverage demonstrations, one common method is transfer learning \cite{JMLR09-taylor}, where one (source) agent is used to speed up learning in a second (target) agent. However, many existing transfer learning methods can provide limited help for complex tasks, since there are assumptions about the source and/or target agent's internal representation, demonstration type, learning method, etc. 

One approach is the Human Agent Transfer~\cite{11AAMAS-HAT-Taylor} (HAT) algorithm, which provided a framework where a source agent could demonstrate policy and a target agent could improve its performance over that policy. As refinement, a Confidence Human Agent Transfer~\cite{2017IJCAI-Wang} algorithm was proposed by leveraging the confidence measurement on the policy. Notice that these methods are different from demonstration learning work like those discussed in~\cite{argall2009survey}, as the target agent is learning to outperform demonstrators rather than just mimic them.

Probabilistic Policy Reuse~\cite{fernandez2006probabilistic} is another transfer learning approach. Like many other existing approaches, it assumes both the source and the target agents share the same internal representations and optimal demonstrations are required. But here we are focusing on improving learning performance without such assumptions. Existed policies could guide the learning direction as shown elsewhere~\cite{da2010using,brys2017multi}, but well-formulated policies could be impracticable due to the complexity of the learning task or the cost of a domain expert's time.

The target agent must handle multiple potential problems.
First, the source agent may be suboptimal. Second, if multiple sources of prior knowledge are considered, they must be combined in a way to handle any inconsistancies.
Third, the source agent typically cannot exhaustively demonstrate over the entire state space; some type of generalization must be used to handle unseen states.
Fourth, the target agent may have a hard time balancing  the usage of the prior knowledge and its own self-learned policy.

In this paper, we introduce \name{} (Dynamic Reuse of Prior), as a interactive method to assist Reinforcement Learning by addressing the above problems. \name{} uses temporal difference models to perform online confidence-based performence measurement on transferred knowledge. In addition, we have three action decision models
to help the target agent balance between following the source advice and following its own learned knowledge. We evaluate \name{} using the domains of Cartpole and Mario, showing improvement over existing methods. Furthermore, through this novel online confidence-based measurement, \name{} is capable of distinguishing the quality of prior knowledge as well as leveraging demonstrations from multiple sources.




\section{Background}
This section presents a selection of relevant techniques. 

\subsection{Reinforcement Learning}

By interacting with an environment, an RL agent can learn a policy to maximize an external reward.
A Markov decision process is common formulation of the RL problem. In a Markov decision process, $A$ is a set of actions an agent can take and $S$ is a set of states. There are two (initially unknown) functions within this process: a transition function
($T:S\times A \mapsto S$)
and a reward function
($R:S\times A \mapsto R$).

The goal of an RL agent is to maximize the expected reward --- different RL algorithms have different ways of approaching this goal. For example, two popular RL algorithms that learn to estimate $Q$, the total long-term discounted reward, are SARSA~\cite{rummery1994line,singh1996reinforcement}:
\[Q(s,a)\leftarrow Q(s,a) + \alpha [r+\gamma Q(s',a')-Q(s,a)]\]
and Q-learning~\cite{watkins1992q}:
\[Q(s,a)\leftarrow Q(s,a) + \alpha [r+\gamma\underset{a'}{max} Q(s',a')-Q(s,a)]\]




\subsection{Human Agent Transfer (HAT)}
The goal of HAT~\cite{11AAMAS-HAT-Taylor} is to leverage demonstration from a source human or source agent, and then improve agents' performance with RL. \textit{Rule transfer}~\cite{taylor2007cross} is used in HAT to remove the requirements on sharing the same internal algorithms representation between source and target agents.
The following steps summarize HAT:

\begin{enumerate}
\item Learn a policy ($ \pi:S \mapsto A$) from the source task.
\item{Train a decision list upon the learned policy as ``IF-ELSE'' rules. }
\item{Bootstrap the target agent's learning with trained decision rules. The target agent's action is guided by rules under a decaying probability. }

\end{enumerate}

\subsection{Confidence Human Agent Transfer (CHAT)}
CHAT~\cite{2017IJCAI-Wang} provides a method based on confidence --- it 
leverages a source agent's/human's demonstration to improve its performance.
CHAT measures the confidence in the source demonstration. Such offline confidence is used to predict how reliable the transferred knowledge is.

To assist RL, CHAT will leverage the source demonstrations to suggest an action in the agent's current state, along with the calculated confidence. For example, CHAT would use Gaussian distribution to predict action from demonstration with a offline probability. If the calculated confidence is higher than a pre-tuned confidence threshold, the agent would consider the prior knowledge reliable and execute the suggested action.

\section{Dynamic Reuse of Prior (\name{})}
This section introduces \name{}, which provides an online confidence-based performance analysis on knowledge transfer to assist reinforcement learning.


Prior research~\cite{chernova2007confidence} used an offline confidence measure of demonstration data, similar to CHAT. In contrast, our approach performs \emph{online confidence-based analysis} on the demonstrations during the target agent's learning process. We introduce two types of temporal difference confidence measurements (section~\ref{sec:tdmodel}) and three types of action decision models (section~\ref{sec:asmodel}),
which differ by whether prior knowledge should be used in the agent's current state.

\name{} follows a three step process:


\begin{enumerate}

\item
Collect a demonstration dataset (state-action pairs).

\item

Use supervised learning to train classifier on the demonstration data. Different types of classifiers could be applied in this step but this paper uses 
a fully connected neural network, and the confidence distribution is calculated through the softmax layer (calculation function in Section~\ref{sec:tdmodel}).

\item
Algorithm~\ref{alg:drop} is used to assist an RL agent in the target task. 
The action decision models will determine whether to reuse the transferred knowledge trained in the previous step or to use the agent's own Q-values. The online confidence model will be updated simultaneously, along with Q-values.

\end{enumerate}

As learning goes on, there will be a balance between using the transferred knowledge and learned Q-values. Notice that we do not directly transfer or copy Q-values in the second step --- the demonstrating agent can be different from the target agent (e.g., a human can teach an agent).
The supervised learning step removes any requirements on the source demonstrator's learning algorithm or representation.  

Relative to other existing work, there are significant advantages of \name{}'s online confidence measurement: First, it removes
%
%
the trial-and-error confidence threshold tuning process.
%
Second,
the target agent's experience is used to measure confidence on demonstrations. \name{}  performs the adaptive confidence-based performance analysis during the target agent's learning. 
This online process can help guarantee the transfer knowledge is adapted to the target tasks.
%
Third, there is no global reuse probability control, a parameter that is 
crucial in other knowledge reuse methods~\cite{2017IJCAI-Wang,11AAMAS-HAT-Taylor,fernandez2006probabilistic} to avoid suboptimal asymptotic performance.
%
%
%

\subsection{Temporal Difference Confidence Analysis}
\label{sec:tdmodel}
The online confidence metric is measured via a temporal difference (TD) approach. For each action source (learned Q function or prior knowledge), we build a TD model to measure the confidence-based performance via experience.

A confidence-based TD model is used to analyze the performance level of every action source with respect to every state. Once an action is taken, the confidence model will update the corresponding action source's confidence value. Generally speaking, an RL agent should prefer the action source with higher confidence level: the expected reward would likely be higher by taking the action from that source.




Our dynamic TD confidence model updates as follows:
\[C(s)\leftarrow (1-F(\alpha)) \times C(s) + F(\alpha) \times [G(r)+\gamma \times C(s')]\]
where $\gamma$ is discount factor, $r$ is reward, and $\alpha$ is the update parameter. For continuous domains, function approximators such as tile coding~\cite{albus1981brains} should be used --- in this work we are using the same discretization approximator as $Q(s,a)$. 
We define two types of knowledge models, described next, although more are possible.

The \textbf{confidence prior knowledge model} is denoted by $CP(s)$. We have 2 update methods: Dynamic Rate Update (DRU) and Dynamic Confidence Update (DCU). For DRU, since \name{} uses a neural network for supervised classification in this paper, we define a dynamic updating rate based on a softmax~\cite[pp.  206--209]{bishop2006pattern} layer's classification distribution:
\[
F(\alpha)= \alpha \times \max \big\{ \frac{1}{\sum_{i} exp(\theta_{i}^T\cdot x)}
\begin{bmatrix}
exp(\theta_{1}^T\cdot x))\\
exp(\theta_{2}^T\cdot x))\\
...\\
exp(\theta_{i}^T\cdot x))
\end{bmatrix}
\big\}
\]
$\theta_{i}$ is the weight vector of the softmax layer and $x$ is the corresponding input. $\max\{\cdot\}$ in the above equation is the output confidence by the network. The update rate of $CP(s)$ will be bounded by the confidence of the corresponding classification. If the confidence is higher, the update rate will be larger (and vice versa). Besides, we use the original reward from the learning task: $G(r)=r$.

For DCU, we use a fixed update rate: $F(\alpha)=\alpha$, but the reward function leverages the confidence:
\[
G(r)= \frac{r}{r\_max}\times \max \big\{ \frac{1}{\sum_{i} exp(\theta_{i}^T\cdot x)}
\begin{bmatrix}
exp(\theta_{1}^T\cdot x))\\
exp(\theta_{2}^T\cdot x))\\
...\\
exp(\theta_{i}^T\cdot x))
\end{bmatrix}
\big\}
\]

In the above equation, $\frac{r}{r\_max}$ is a normalized reward ($r\_max$ denotes the maximum absolute reward value) and G(r) re-scales the reward using confidence distribution.



The \textbf{confidence Q knowledge model} is denoted by $CQ(s)$. $CQ(s)$ uses the same update methods with $F(\alpha)=\alpha$ and $G(r)=r$. $CQ(s)$ will be updated only if an action is provided through $Q(s,a)$.


\subsection{Action Selection Methods}
\label{sec:asmodel}
Given these TD-based confidence models, we introduce three action selection methods that balance an agent's learned knowledge (CQ) with its prior knowledge (CP).


The {\bf hard decision model} (HD) is greedy and attempts to maximize the current confidence expectation. Given current state $s$, action source $AS$ is selected as:
\[
AS= {\arg\max}[\{CQ(s),CP(s)\}],
\]
where ties are broken randomly.



The {\bf soft decision model} (SD) decides action source using probability distribution. To calculate the decision probability, we first normalize $CQ(s)$ and $CP(s)$: $R= \max \{|CQ(s)|,|PQ(s)|\}$, $rCQ=CQ(s)/R $, $rCP=CP(s) / R$. Then rescale $rCQ$ and $rCP$ using the hyperbolic tangent function (using $rCQ$ as example):
\[
\tanh(rCQ)=\frac{e^{rCQ}-e^{-rCQ}}{e^{rCQ}+e^{-rCQ}} 
\]
The probability of selecting action source is defined as:

\vspace{-5pt}
\begin{equation} \label{eq:qpmodel}
AS  ={
  \begin{cases}
  Q   &  P=\frac{\tanh \left({rCQ} \right) +1}{\tanh \left( rCP \right) +\tanh \left({rCQ} \right) +2}  \\
  Prior   & P=\frac{\tanh \left({rCP} \right) +1}{\tanh \left( rCP \right) +\tanh \left({rCQ} \right) +2} 
  \end{cases}
  }
\end{equation}
\vspace{-5pt}

If the confidence in the prior knowledge is high, the target agent would follow the prior with high probability. If the confidence in the prior knowledge is low, it might still be worth trying, but with lower probability. If the confidence in the prior knowledge is very low, the probability would then be almost zero. 


The third model is the {\bf soft-hard-$\epsilon$ decision model} (S-H-$\epsilon$), shown in Algorithm~\ref{alg:shed}. This method takes advantage of the above two models by adding an $\epsilon$-greedy switch. That is to say, we have added an $\epsilon$-greedy policy over HD and SD: S-H-$\epsilon$ can both greedily exploit the confidence value and also perform probabilistic exploration. Notice that our method could also handle multiple-source demonstrations. By adding parallel prior models, the above $AS$ (in Equation~\ref{eq:qpmodel}) could be expanded into multiple cases: 
\vspace{-5pt}
\begin{equation} \label{eq:pmodels}
AS  = {
  \begin{cases}
  {Prior}_1  &  P_1=\frac{\tanh({rCP}_1) +1}{\underset{i}{\sum}\{\tanh({rCP}_i)+1\}}  \\
  {Prior}_2   &   P_2=\frac{\tanh({rCP}_2) +1}{\underset{i}{\sum}\{\tanh({rCP}_i)+1\}}  \\
  \hspace{10pt}... &\hspace{45pt}...\\
  {Prior}_i   &  P_i=\frac{\tanh({rCP}_i) +1}{\underset{i}{\sum}\{\tanh({rCP}_i)+1\}}\end{cases}
  }
\end{equation}
\vspace{-13pt}


\vspace{-5pt}
\begin{algorithm2e}[tbh]
\KwIn{Prior knowledge model PM }

\For{each episode}{
Initialize state $s$ to start state\\
\For{each step of an episode}{
  
  \eIf{rand() $\leq \epsilon$}
  {\%Exploration:\\
  $a \gets$ random action
  }
  {\%Action source (\textit{AS}) selected via HD, SD, or S-H-$\epsilon$:\\ $AS \gets $ Action Decision Model\\
    \eIf{$AS ==$ Prior Knowledge}
    {$a \gets$ action from Prior Knowledge\\
    Update CP}
    {$a \gets$ action that maximizes $Q$\\
    Update CQ}
  }
  Execute action $a$\\
  Observe new state $s^\prime$ and reward $r$\\
  Update $Q$ (SARSA, Q-Learning, etc.);\label{Line:Qupdate}\\
  }
}
 
 \caption{\name: Target Learning Bootstrap}
 \label{alg:drop}
\end{algorithm2e}

\vspace{-10pt}


\subsection{Optimum Convergence Property}
Here we discuss the theoretical analysis of the convergence of \name{}.
$\pi_{P}$ and $\pi_{Q}$ denote the policy of prior knowledge and learned Q knowledge, respectively. 
Given a fixed policy, the optimal convergence of TD iteration is proven, as was done by~\cite{sutton1988learning}. For the static prior knowledge policy, we have $\mathds{E}[CP(s)] = CP^*(s)$. For the Q knowledge, $CQ(s) = \sum\limits_{a}\pi_{Q}(a|s)q(s,a)$. Since $q(s,a)$ is updated independently (Line~\ref{Line:Qupdate} of Algorithm~\ref{alg:drop}) and the Q-learning's convergence is guaranteed by~\cite{melo2001convergence}, we also have $\mathds{E}[CQ(s)] = CQ^*(s)$ on the converged $\pi^*_{Q}$. We will then prove that whatever the quality of prior knowledge is, \name{} will not harm Q-learning's asymptotic performance.

\begin{proof}
Given state $s$, if $CP^*(s) \geq CQ^*(s)$, which means the optimal $\pi^*_{P}(s)$ is better than $\pi^*_{Q}(s)$, the proof is trivial because following prior knowledge would result in higher reward. On states where $CP^*(s) < CQ^*(s)$, according to Line~\ref{Line:prob} of Algorithm~\ref{alg:shed} the probability of using suboptimal action is $\epsilon\times\frac{\tanh \left({rCP} \right) +1}{\tanh \left( rCP \right) +\tanh \left({rCQ} \right) +2} < \epsilon$, which means the suboptimal action is under $\epsilon$-greedy control.
\end{proof}

We therefore conclude that \name{} should guarantee the learning optimum, and Q-learning's convergence will not be harmed even if the prior knowledge contains suboptimal data.



\begin{algorithm2e}[t!]
\KwIn{$CQ, CP, $ State $s$}
$R= \max \{|CQ(s)|,|PQ(s)|\}$\\
$rCQ=CQ(s) / R $  \\
$rCP=CP(s) / R $  \\
 \eIf{rand() $\leq \epsilon$}
    {\eIf{rand() $\leq \frac{\tanh \left({rCQ} \right) +1}{\tanh \left( rCP \right) +\tanh \left({rCQ} \right) +2}$\label{Line:prob}}
  {
   $AS$ = Prior\ Knowledge
  }
  {
   $AS$ = Q\ \ Knowledge
  }
    }
    {
  $AS= {\arg\max}[\{CQ(s),CP(s)\}]$}
 $\Return\ AS $ \%Action source
 \caption{S-H-$\epsilon$ : Hard-Soft-$\epsilon$ Decision Model}
 \label{alg:shed}
\end{algorithm2e}

\section{Experiment Setup}
This section details our 
experimental methodology.

\subsection{Experiment Domains}
We evaluate our method in two domains: Cartpole and Mario.

{\bf Cartpole} is a classic control problem -- balancing a light-weight pole hinged to a cart. Our Cartpole simulation is based on the open-source OpenAI Gym~\cite{brockman2016openai}. This task has a continuous state space; the world state is represented as 4-tuple vector: position of the cart, angle of the pole, and their corresponding velocity variables. The system is controlled by applying a force of +1 or -1 to the cart. Cartpole's reward function is designed as: $+1$ for every surviving step and $-500$ if the pole falls.

{\bf Mario} is a benchmark domain~\cite{karakovskiy2012mario} based on Nintendo's Mario Brothers. We train the Mario agent to score as many points as possible. To guarantee the diversity and complexity of tasks, our simulation world is randomly sampled from a group of one million worlds. The world state is represented as a 27-tuple vector, encoding the agent's state/position information, surrounding blocks, and enemies~\cite{suay2016learning}.
There are $12 \ ( 3\times2 \times 2)$ possible actions (move direction $\times$ jump button $\times$ Run/Fire button).





\subsection{Methodology}
\name{} can work with demonstrations collected from both humans and other agents. In our experiments, demonstrations are collected either from a human participant (one of the authors of this paper) via a simulation visualizer, or directly from an agent executing the task.


We use a ``4-15-15-2'' network (15 nodes in two hidden layers) network in Cartpole and a ``27-50-50-12'' network in Mario. To benchmark against CHAT, we use the same networks as the confidence models used by \name{}. To benchmark against HAT, J48 \cite{Quinlan1993} is used to train decision rules. Our classifiers are trained using classification libraries provided by Weka 3.8~\cite{weka}. For both CHAT and HAT, the self-decaying reuse probability control parameter $\Phi$ was tuned to be 0.999 in Cartpole and 0.9999 in Mario.
Target agents in both Cartpole and Mario are using Q-learning algorithm. In Cartpole, we use $\alpha = 0.2, \gamma = 0.9, \epsilon = 0.1$. In Mario, we use $\alpha = \frac{1}{10 \times 32}, \gamma = 0.9, \epsilon = 0.1$. These parameters are set to be consistent with previous research~\cite{2017IJCAI-Wang,brys2015policy} in these domains. 

Experiments are evaluated in terms of learning curves, the jumpstart, the total reward, and the final reward. Jumpstart is defined as the initial performance improvement, compared to an RL agent with no prior knowledge. The total reward accumulates scores every 5 percent of the whole training time. Experiments are averaged over 10 learning trials and t-tests are performed to evaluate the significance. Error bars on the learning curves show the standard deviation.\footnote{Our code and demonstration data will be made available after acceptance.}





\section{Experimental Results}
This section will present and discuss our main experimental results. We first show the improvement over existing knowledge reuse algorithms, HAT and CHAT, as well as baseline learning. Then we show \name{} is capable of leveraging different quality demonstrations from multiple sources. Finally we will evaluate how \name{} could be used for interactive RL by efficiently involving a human demonstrator in the loop.\footnote{Due to length limitation, extra results are shown in the anonymous link for review: http://dropextra.webs.com/}

\subsection{Improvement over Baselines}

In Cartpole, we first let a trained agent demonstrate 20 episodes (average number of steps: 821 $\pm$ 105) and record those state-action pairs. In Mario, we let a trained agent 
demonstrate 20 episodes (average reward: 1512 $\pm$ 217). 

\name{} is then used with these demonstration datasets. As benchmarks, we run HAT and CHAT on the same datasets, and Q-learning is run without prior knowledge. Learning performance is compared in Table~\ref{table:compall}. \name{} with different models outperforms the baselines. The top two scores for each type of performance are underlined. \name{} with DRU and S-H-$\epsilon$ model has achieved the best learning result and further discussions in the next
sections use this setting. Statistically significant ($p<10^{-4}$) improved scores in Table~\ref{table:compall} are in bold.  
There is no significant difference ($p>0.05$) from CHAT and HAT, for the final reward of Mario.

\begin{figure}[t!]
\centering
\centerline{\includegraphics[width=0.95\columnwidth]{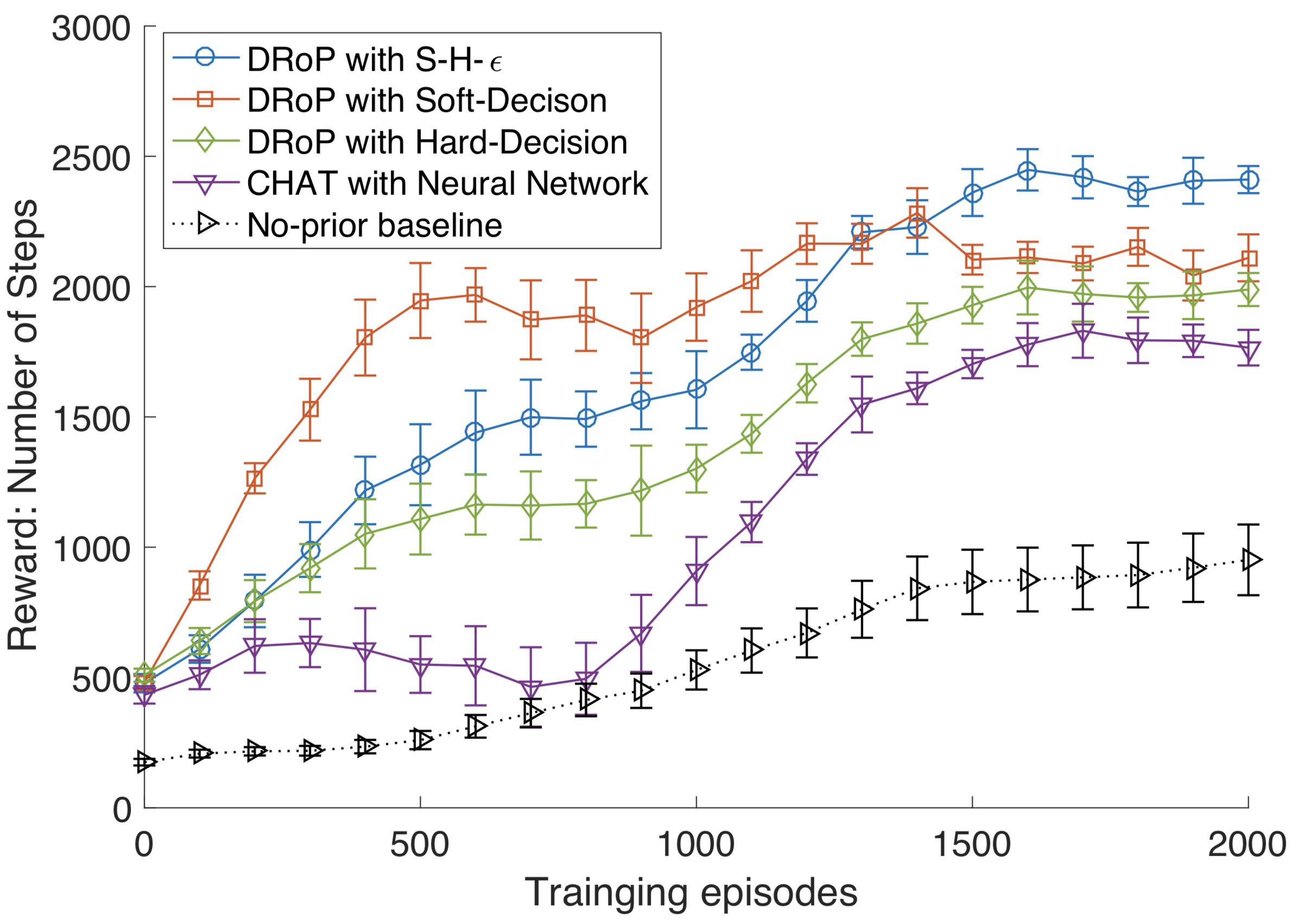}}
\vspace{-10pt}
\caption{Cartpole learning curves}
\vspace{-5pt}
\label{fig:cartpole_score}
\end{figure}

\begin{figure}[t!]
\centering
\centerline{\includegraphics[width=0.95\columnwidth]{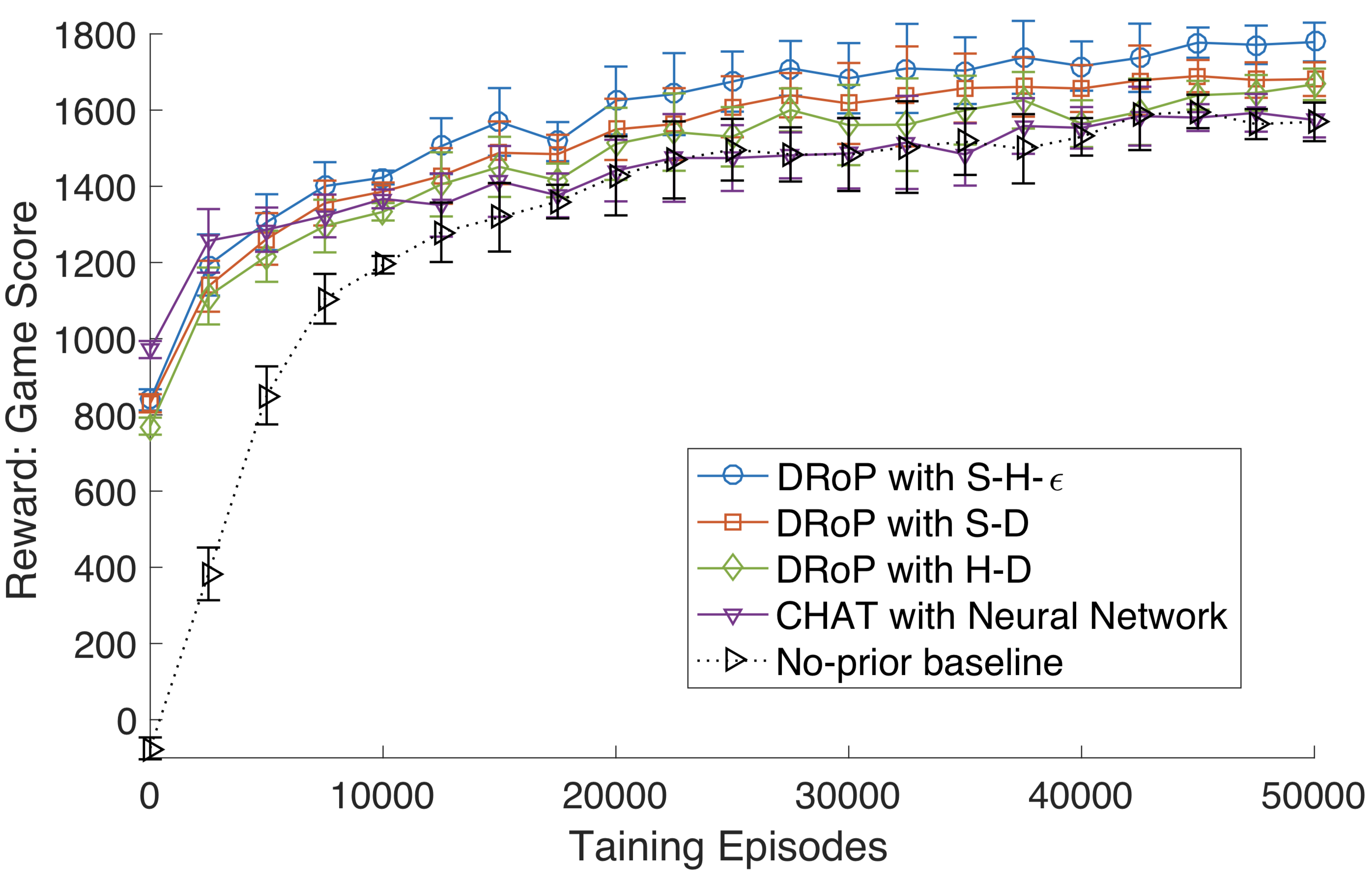}}
\vspace{-10pt}
\caption{Mario learning curves}
\vspace{-8pt}
\label{fig:mario_score}
\end{figure}

To highlight the improvement, Figure~\ref{fig:cartpole_score} and~\ref{fig:mario_score} show the learning curves of \name{} using DRU method. All three action selection schemes of \name{} (DRU) outperform HAT, CHAT, and baseline learning, indicating that \name{} 
is most effective. 

\begin{table*}[tbh]
\centering

\begin{tabular}{||c |c |c |c |c| c|c ||} 
 \hline
\multirow{2}{*}{Method} & \multicolumn{3}{c|}{Cartpole} & \multicolumn{3}{c||}{Mario} \\ \cline{2-7} 
                   &Jumpstart& Total Reward  &Final Reward    &jumpstart & Total Reward &Final Reward\\ \hline \hline
Q-Learning         &N/A   &11653     &951 $\pm$ 36    &N/A   &27141     &1569 $\pm$ 51      \\ 
HAT                &\textbf{225}   &\textbf{16283}     &\textbf{1349 $\pm$ 84}   &\textbf{651}   &25223     &1577 $\pm$ 49      \\ 
CHAT               &\textbf{258}   &\textbf{22692}     &\textbf{1766 $\pm$ 68}   &\underline{\textbf{1046}}  &\textbf{30144}     & 1574 $\pm$ 46     \\ 
DCU, H-D           &\textbf{298}   &\textbf{29878}     &\textbf{1994 $\pm$ 62}   &\textbf{829}   &\textbf{31021}     &\textbf{1675 $\pm$ 59}      \\ 
DCU, S-D           &\textbf{301}   &\textbf{33498}     &\textbf{2085 $\pm$ 79}   &\textbf{880}   &\textbf{31436}     &\textbf{1690 $\pm$ 62 }     \\ 
DCU, S-H-$\epsilon$& \textbf{308}        &\textbf{35312}  &\textbf{\underline{2383 $\pm$ 71}}   &\textbf{909} &\textbf{\underline{32108}} &\textbf{\underline{1752 $\pm$ 55}}      \\
DRU, H-D           &\underline{\textbf{334}}   & \textbf{29563}    &\textbf{1989 $\pm$ 63}   &\textbf{845}   &\textbf{30644}     &\textbf{1668 $\pm$ 41 }     \\ 
DRU, S-D           &\underline{\textbf{305}}   &\underline{\textbf{38576}}     &\textbf{2111 $\pm$ 90}   &\textbf{905}   &\textbf{31690}     &\textbf{1681 $\pm$ 44 }     \\ 
DRU, S-H-$\epsilon$&\textbf{303} &\underline{\textbf{35544}}       &\underline{\textbf{2411 $\pm$ 56}}   &\underline{\textbf{915}} &\underline{\textbf{33022}}       &\textbf{\underline{1779 $\pm$ 61}}      \\ \hline

\end{tabular}
\vspace{-6pt}
\caption{This table compares baselines (methods 1 to 3) with  \name{} using different models (methods 4 to 9). Jumpstart, total reward, and final reward are shown. The top two scores of each column are underscored and significant improvements over Q-learning are bold.}
\vspace{-10pt}
\label{table:compall}
\end{table*}


\subsection{\name{}ping Low-quality Demonstrations}
We consider using suboptimal demonstrations to see how well the online confidence-based analysis mechanism can handle poor data without harming the optimal convergence. Here we have five different groups of demonstrations (recorded from different agents), ranging from completely random to high performing (shown in Tables~\ref{table:cartpoledemos} and~\ref{table:mariodemos}).

We first evaluate our method individually with the five demonstration datasets. Cartpole results are shown in Table~\ref{table:cartpoledemos} and Mario results are shown in Table~\ref{table:mariodemos}. As we can see, the quality of the demonstration does effect performance, and better demonstrations lead to better performance. However, what is more important is whether poor demonstrations hurt learning. If we look at the results of using randomly generated demonstrations, we find that even if the jumpstart is negative (i.e., the initial performance is hurt by using poor demonstrations), the final converged performance is almost the same as learning without the poor demonstrations. In addition, the converged reuse frequency (average percentage of actions using the prior knowledge) of random demonstration is almost zero, which means the \name{} agent has learned to ignore the poor demonstrations. As the demonstration quality goes higher (from L1 to L4), \name{} will reuse the prior knowledge with higher probability.

We evaluate the multiple-case model (Equation~\ref{eq:pmodels} by providing the above demonstrations simultaneously to \name{} and results are shown in Table~\ref{table:cartpoledemogroups}. When low-quality demonstrations are mixed in the group, we see a decreased jumpstart from both CHAT and \name{}, relative to that seen in Table~\ref{table:cartpoledemos}. In contrast, \name{} distinctly reuses the high-quality data more often and achieves better performance.

\begin{table}[tb]
\centering
\begin{tabular}{||c|c|c|c||} 
\hline
 \shortstack{Demo\vspace{-2pt}\\Performance}& \shortstack{Jump-\vspace{-2pt}\\start} & \shortstack{Converged \vspace{-2pt}\\ Performance} & \shortstack{Converged\vspace{-2pt}\\  Reuse\vspace{-2pt} \\Frequency}\\ 
\hline \hline 

Q-Learning     & N/A &951 $\pm$ 136 & N/A\\
\hline 
Rand:   15 $\pm$ 7    &-5 &942 $\pm$ 142&0.02 $\pm$ 0.01 \\
L1:     217 $\pm$ 86  &153  &1453 $\pm$ 96 &0.12 $\pm$ 0.03  \\ 
L2:     435 $\pm$ 83  &211 &1765 $\pm$ 112 &0.17 $\pm$ 0.04 \\ 
L3:     613 $\pm$ 96  &278 &2080 $\pm$ 86 &0.21 $\pm$ 0.02 \\
L4:     821 $\pm$ 105 &303 &2411 $\pm$ 56 &0.32 $\pm$ 0.03
\\

 \hline
\end{tabular}
\vspace{-6pt}
\caption{This table shows the performance of Q-learning and \name{} (DRU, S-H-$\epsilon$) upon 5 different levels of demonstrations in Cartpole.}
\vspace{-5pt}
\label{table:cartpoledemos}
\end{table}

\begin{table}[tpb]
\centering

\begin{tabular}{||c |c |c |c ||} 
\hline
 \shortstack{Demo\vspace{-2pt}\\Performance}&\shortstack{Jump-\vspace{-2pt}\\start} & \shortstack{Converged \vspace{-2pt}\\ Performance} & \shortstack{Converged\vspace{-2pt}\\  Reuse \vspace{-2pt}\\ Frequency}\\ 
\hline \hline 
 
 Q-Learning& N/A &1569 $\pm$ 51 & N/A\\
 \hline
 Rand: -245 $\pm$ 11&-52 &1552 $\pm$ 72 & 0.01 $\pm$ 0.01\\ 
 L1: 315 $\pm$ 183&336 &1582 $\pm$ 67 & 0.08 $\pm$ 0.02 \\ 
 L2: 761 $\pm$ 195&512 &1601 $\pm$ 73 &0.15 $\pm$ 0.05 \\ 
 L3: 1102 $\pm$ 225&784 &1695 $\pm$ 81 &0.19 $\pm$ 0.03 \\
 L4: 1512  $\pm$ 217&906 & 1779 $\pm$ 61 &0.28 $\pm$ 0.04
 \\

 \hline
\end{tabular}
\vspace{-6pt}
\caption{This table shows the performance of Q-learning and \name{} (DRU, S-H-$\epsilon$) upon 5 different levels of demonstrations in  Mario.}
\label{table:mariodemos}
\vspace{-6pt}
\end{table}

\begin{table}[t!]
\centering

\begin{tabular}{||c |c |c|c||} 
\hline
Method & Jumpstart & \shortstack{Converged \vspace{-2pt}\\ Performance}   & \shortstack{Converged\vspace{-2pt}\\  Reuse Frequency} \\ 
\hline \hline 

CHAT & 191 & 983 $\pm$ 151 & 0.05 $\pm$ 0.02
\\ 
\hline

\multirow{5}{2.4em}{\name{}}&\multirow{5}{1.7em}{253} & \multirow{5}{4.3em}{2286 $\pm$ 91}&Rand: 0.02 $\pm$ 0.01 \\
      & & &L1: 0.05 $\pm$ 0.01  \\ 
     & & &L2: 0.06 $\pm$ 0.02 \\ 
       & & &L3: 0.11 $\pm$ 0.03 \\
    & & &L4: 0.23 $\pm$ 0.02\\

 \hline
\end{tabular}
\vspace{-6pt}
\caption{This table shows the performance of \name{} (DRU, S-H-$\epsilon$) and CHAT upon multiple sources of demonstrations in Cartpole.}
\label{table:cartpoledemogroups}
\vspace{-8pt}
\end{table}

\subsection{\name{}-in Requests for Demonstrations}


We have shown that \name{} is capable of analyzing the quality of demonstration. This section asks a different question --- can \name{} use these confidence values to productively request additional demonstrations from a human or agent? 

In Mario, we first recorded 20 episodes of demonstrations from an human expert with an average score of 1735. We then used \name{} to assist an RL agent's learning. After a short period of training (1000 episodes), we then use the following steps to ask for additional demonstrations from the same human demonstrator over in the next 20 episodes:

\begin{enumerate}
\item Calculate average confidence of prior knowledge (i.e., $CP(s)$) at each step of the current episode:
\vspace{-5pt}
\[ AveC=\frac{1}{steps}\times\underset{i}{\sum}CP(s_i)\]
\vspace{-15pt}
\item Use a sliding window of 10 $\times$ 10 to scan neighbourhood positions and calculate the average ``$CP(s)$'' within that sliding window. 
\vspace{-1pt}
\item If the averaged CP value is smaller than $AveC$, request a demonstration of 20 actions, starting at the current state.
\vspace{-12pt}
\item Add the above recorded state-action pairs into the request demonstration dataset of \name{}. 
\vspace{-3pt}
\end{enumerate}
The requested demonstration dataset is still recorded within 20 episodes, but the time spent actively demonstrating is reduced by 44\%, relative to demonstrating for 20 episodes (shown in Table~\ref{table:humandemo}), because demonstrations are requested only when the agent's confidence of prior knowledge is low. The time cost of demonstration collection is only 2\% of the baseline training time, highlighting the efficincy of \name{}. We then compare it with the originally collected demonstration from the same human. 

\begin{figure}[t!]
\centering
\centerline{\includegraphics[width=0.95\columnwidth]{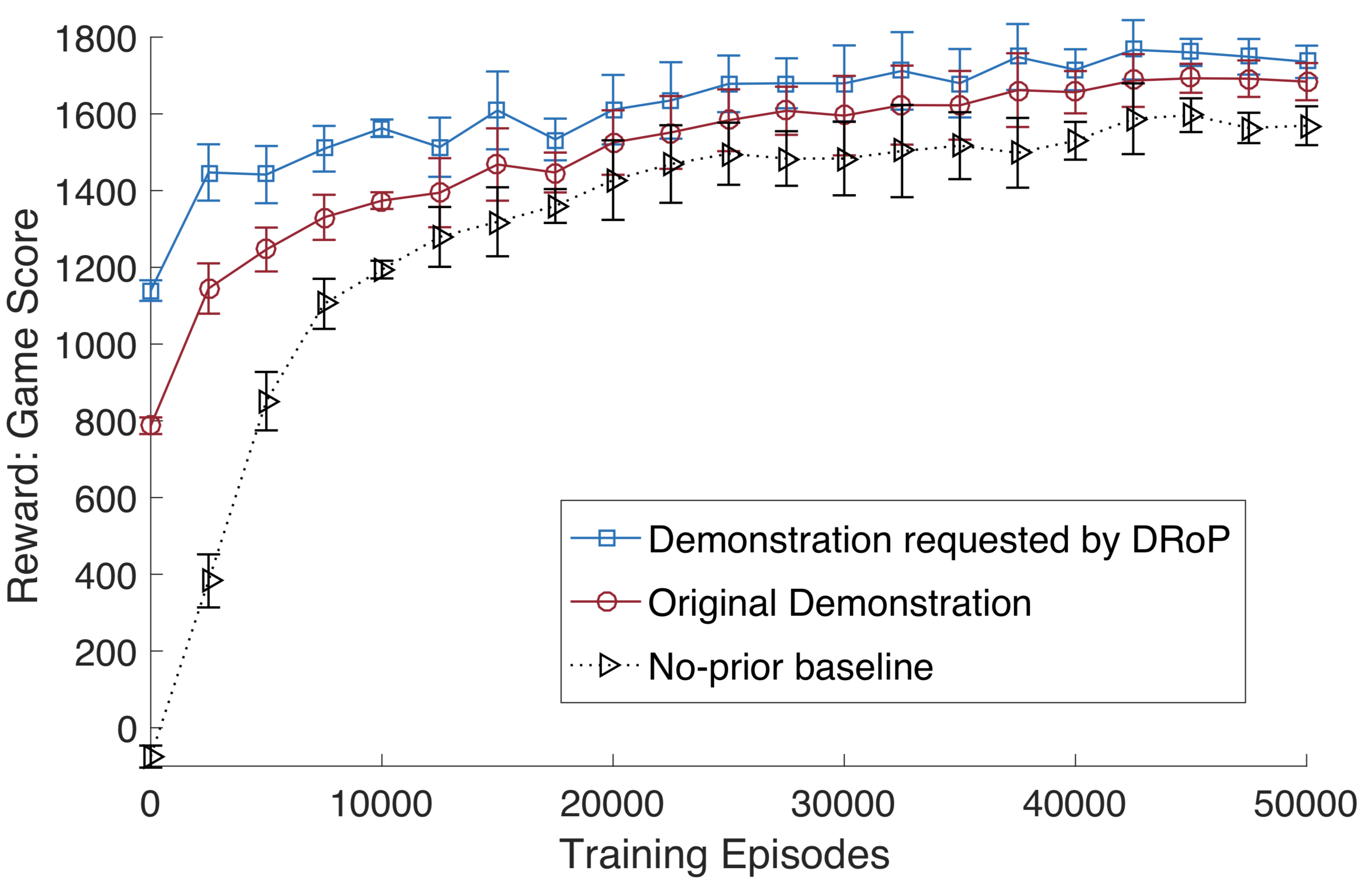}}
\vspace{-5pt}
\caption{Mario learning curves using demonstration requested by \name{} and original demonstration from human expert.
}
\vspace{-5pt}
\label{fig:mario_human}

\end{figure}

Figure~\ref{fig:mario_human} shows the performance comparison between the two demonstration datasets: 20 episodes of original human demonstrations and 20 episodes requested by \name{}. Notice that even though human's demonstration performance is higher than the L4 dataset from the previous section, the actual jumpstart of the former is instead lower. This is potential evidence that a virtual agent could not ``digest'' the entire human demonstrator's knowledge. In contrast, learning improvement from the extra demonstration requested by \name{} is higher. \name{} would request the demonstration from human only in states where the knowledge confidence is relatively low. Therefore, we know that the target agent truly needs these requested demonstrations. \name{} improved the overall learning effectiveness by requesting less, but critical, demonstration data.


\begin{table}[tpb]
\centering
\begin{tabular}{||c |c |c |c  ||} 
 \hline
 Souce &  Time Cost & Jumpstart & \shortstack{Converged \vspace{-2pt}\\ Performance} 
 \\ [0.5ex] 
 \hline \hline 
 Baseline    & 15325 s & N/A &951 $\pm$ 136 \\
\hline 
 Original &623 s  &862 &1684 $\pm$ 49 \\ 
 Request  &348 s &1214 &1736 $\pm$ 42   
 \\

 \hline
\end{tabular}
\vspace{-5pt}
\caption{This table compares the original human demonstration and demonstration requested by \name{}  (DRU, S-H-$\epsilon$) in Mario.}
\vspace{-8pt}
\label{table:humandemo}
\end{table}


\section{Conclusion and Future Work}
This paper has introduced \name{} and evaluated it in two domains. This work shows that by integrating offline confidence with online temporal difference analysis, knowledge transfer from source agents or humans can be successfully achieved. \name{} outperformed both learning without prior knowledge and recent transfer methods.

\name{}'s confidence measurement is based on temporal difference (TD) models. Results suggest that such online confidence techniques can provide reasonable and reliable analysis of the quality of prior knowledge.

Two temporal difference methods and three action selection models are introduced in this work. It is shown that \name{}'s decision mechanism can leverage multiple sources of demonstrations. 
In our experimental domains, DRU with S-H-$\epsilon$ produced the best performance.

Results have shown that demonstrations requested by \name{} can significantly improve the RL agent's learning process, leading to a more efficient collaboration between two very different types of knowledge entities: humans and virtual agents.

There are a number of interesting directions for future work, including the following.
First, we will explore model-based demonstrations to see if any particular model structure could provide better confidence measurement than classification equally over all state-action pairs. 
Second, we will use \name{} to build a lifelong online learning system. Our current work could analyze and selectively reuse transferred static prior knowledge and the goal is to let the learning system automatically refine that knowledge model during learning.



\clearpage

\bibliographystyle{named}
\bibliography{refs}

\end{document}